\documentclass{article}
\usepackage{spconf,amsmath,graphicx,amssymb,color,threeparttable}
\usepackage[ruled]{algorithm2e}
\usepackage[hidelinks]{hyperref}
\usepackage{subcaption}
\usepackage{bm}
\usepackage{amsthm,amsmath,amssymb}
 
\usepackage{mathrsfs}
\usepackage[OT1]{fontenc}

\usepackage{booktabs}

\def\red#1{\textcolor{red}{#1}}
\newcommand{\tabincell}[2]{\begin{tabular}{@{}c#1@{}}#2\end{tabular}}

\title{Backdoor Defense via Suppressing Model Shortcuts}

\name{Sheng Yang$^{1}$ \qquad Yiming Li$^{1}$\thanks{Corresponding author: Yiming Li (\href{mailto:li-ym18@mails.tsinghua.edu.cn}{li-ym18@mails.tsinghua.edu.cn}). } 
\qquad Yong Jiang$^{1,2}$ \qquad Shu-Tao Xia$^{1,2}$}

            \address{$^{1}$Tsinghua Shenzhen International Graduate School, Tsinghua University\\
            $^{2}$Research Center of Artificial Intelligence, Peng Cheng Laboratory \\
                  \{yangs22, li-ym18\}@mails.tsinghua.edu.cn;
                \{jiangy, xiast\}@sz.tsinghua.edu.cn}

\begin{document}
%
\maketitle
\begin{abstract}
Recent studies have demonstrated that deep neural networks (DNNs) are vulnerable to backdoor attacks during the training process. Specifically, the adversaries intend to embed hidden backdoors in DNNs so that malicious model predictions can be activated through pre-defined trigger patterns. In this paper, we explore the backdoor mechanism from the angle of the model structure. We select the skip connection for discussions, inspired by the understanding that it helps the learning of model `shortcuts' where backdoor triggers are usually easier to be learned. Specifically, we demonstrate that the attack success rate (ASR) decreases significantly when reducing the outputs of some key skip connections. Based on this observation, we design a simple yet effective backdoor removal method by suppressing the skip connections in critical layers selected by our method. We also implement fine-tuning on these layers to recover high benign accuracy and to further reduce ASR. Extensive experiments on benchmark datasets verify the effectiveness of our method.
\end{abstract}
\section{Introduction}
\label{sec:intro}

With high effectiveness and efficiency, deep neural networks (DNNs) are widely used in many areas, such as face recognition \cite{tang2004video,wang2018cosface,qiu2021end2end}. In general, training a well-performed DNN usually needs a lot of training samples and computational resources. Therefore, third-party resources ($e.g.$, third-party training data) are usually involved in the training process.

However, recent studies demonstrated that using third-party resources also introduced a new security threat. This threat was called backdoor attack \cite{gu2019badnets,li2022few,qi2023revisiting}. Different from adversarial attacks \cite{bai2020targeted,liu2022watermark,gu2022segpgd}, the backdoor adversaries can modify some benign training samples or directly control the training loss to implant malicious model behaviors. The backdoor is a latent connection between adversary-specified trigger patterns and the target label. In the inference process, the adversaries can activate embedded backdoors with trigger patterns. Backdoor attacks are stealthy since attacked models behave normally on predicting benign data, posing huge threats to the applications of DNNs \cite{li2022backdoor}.



Currently, there are many defenses to reduce backdoor threats. For example, model-repairing-based defenses \cite{liu2018fine} intended to directly remove backdoors of attacked models; Detection-based methods \cite{li2022test,guo2021aeva,guo2023scale} identified whether the suspicious samples or models are backdoored; Pre-processing-based ones \cite{li2021backdoor} perturbed input samples before feeding them into the deployed model for predictions. We noticed that almost all existing defenses treated different model structures equally. At most, they treat all models as two separate parts, including the feature extractor and the fully-connected layers, and analyzed backdoor behaviors in the feature space. It raises an intriguing and important question: \emph{Are backdoor attacks independent of model structure}?



In this paper, we explore the backdoor mechanism from the aspect of the model structure. We select one of the most successful and widely adopted structure components, $i.e.$, the skip connection \cite{he2016deep} for the analysis. We believe that this structure is highly correlated to backdoors since it helps the learning of `shortcuts' \cite{wu2020skip} where backdoor triggers are usually easier to be learned. Specifically, we gradually reduce the outputs of skip connection (by multiplying a parameter smaller than 1) in different layers of backdoored models to study their effects. We observe that the attack success rate (ASR) drops significantly more quickly than benign accuracy (BA), especially on critical layers. Motivated by this finding, we propose a simple yet effective backdoor defense by suppressing the skip connections in critical layers selected by our method. To reduce the side effects of shortcut suppression, we also implement fine-tuning on these layers to recover high BA and to further reduce ASR.

In conclusion, our main contributions are three-fold: 
\textbf{1)} We reveal that backdoor threats have an underlying correlation to the skip connection. To the best of our knowledge, this is the first attempt trying to analyze backdoor mechanisms from the aspect of model structure. \textbf{2)} Based on our observations, we propose a simple yet effective backdoor defense, dubbed shortcut suppression with fine-tuning (SSFT). \textbf{3)} We conduct extensive experiments on benchmark datasets, verifying the effectiveness of our defense method.


\begin{figure*}[ht]
    \centering
    \includegraphics[width=0.9\textwidth]{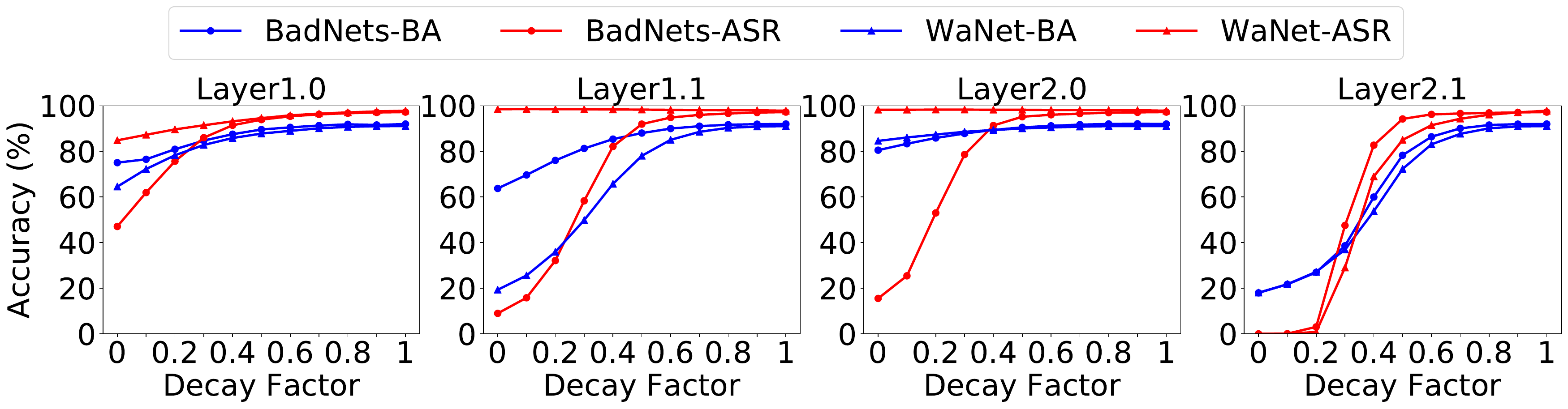}
    \vspace{-0.8em}
    \end{figure*}
\begin{figure*}[ht]
    \centering
    \includegraphics[width=0.9\textwidth]{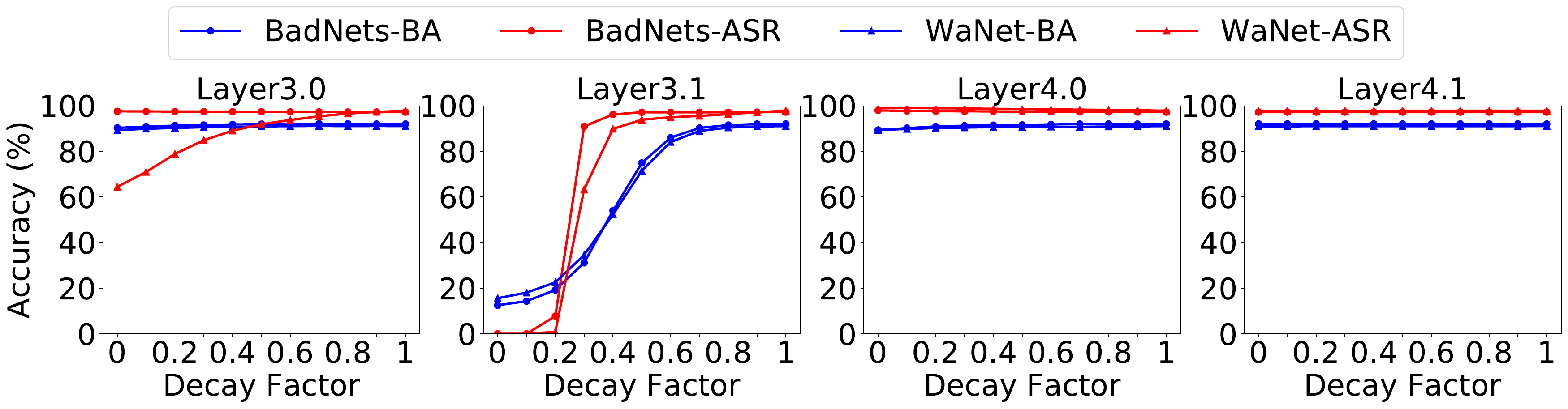}
    \vspace{-0.6em}
    \caption{The benign accuracy (BA) and attack success rate (ASR) $w.r.t.$ the introduced decay factor $\gamma$ of skip connections in different layers of models attacked by BadNets and WaNet on the CIFAR10 dataset.}
    \label{fig:experiments in inference stage}
\end{figure*}

\section{The Effects of Skip Connections}
\label{sec:motivation}


In this section, we analyze how skip connections influence backdoor attacks from both training and inference stages. We adopt ResNet-18 \cite{he2016deep} on the CIFAR-10 dataset as an example for our discussions, as follows:


\vspace{0.3em}
\noindent \textbf{Training Stage.}
We introduce a decay factor $\gamma \in [0,1]$ to reduce gradients from the skip connections to suppress their effects. Our goal is to explore whether the suppression can alleviate the backdoor attack. The gradient suppression is conducted on all skip connections, as denoted in E.q.(\ref{eq:1}).

\begin{equation}
\begin{aligned}
\bm{z_{i}}&=f(\bm{z_{i-1}})+\gamma \cdot \bm{z_{i-1}} \\
\Rightarrow \frac{\partial \bm{z_{i}}}{\partial \bm{w_{i-1}} }&=\frac{\partial f(\bm{z_{i-1}})}{\partial \bm{z_{i-1}}} \cdot \frac{\partial \bm{z_{i-1}}}{\partial \bm{w_{i-1}}}+\gamma   \cdot \frac{\partial \bm{z_{i-1}}}{\partial \bm{w_{i-1}}},
\label{eq:1}
\end{aligned}
\end{equation}
where $\bm{z_{i}}$ is the output of $i$-th layer, $\bm{w_{i}}$ is its weights, and $f$ represents the convolution part of the layer.

However, we find that assigning small $\gamma$ has almost no influence on attack success rate (ASR) and benign accuracy (BA). It indicates that backdoors can also be embedded into structures other than skip connections. It is consistent with the fact that existing methods are effective in attacking DNNs having no skip connection ($e.g.$, VGG \cite{simonyan2015very}).


\vspace{0.3em}
\noindent \textbf{Inference Stage.}
Similar to the procedures in the training stage, we also introduce the decay factor $\gamma$ in skip connections during the inference process. Specifically, we gradually reduce $\gamma$ in each layer of models attacked by BadNets \cite{gu2019badnets} and WaNet \cite{nguyen2021wanet} to observe its influence on BA and ASR. As shown in Figure \ref{fig:experiments in inference stage}, we have three key observations. Firstly, the attack success rate (ASR) may drop more quickly than benign accuracy (BA), especially on critical layers ($e.g.$, layer 2.1\&3.1). Secondly, critical layers do not necessarily contain the first or the last layer that was typically used in existing defenses \cite{tran2018spectral,hayase2021spectre,huang2022backdoor}. Thirdly, different attacks correlate to similar critical layers, which are most probably related to the model structure and training data.

\section{The Proposed Method}

\subsection{Preliminaries}
\noindent \textbf{Threat Model.} 
In this paper, we consider the scenarios that defenders ($i.e.$, model users) obtain a third-party model that could be backdoored. They have no information about the attack while having a few local benign samples.  


\vspace{0.3em}
\noindent \textbf{Defender's Goals.} 
Defenders intend to have low attack success rate on poisoned testing samples while having high benign accuracy on predicting benign testing samples.


\begin{table*}[!t]
\centering
\caption{The main results (\%) on the CIFAR-10 and the GTSRB dataset. We mark failed cases (ASR$>80\%$ or BA $<80\%$) in red and the best result among all defenses in boldface. Underlined values are the second-best results.}
\vspace{-0.8em}
\label{table:experiments results}
\begin{threeparttable}
\scalebox{0.76}{
\begin{tabular}{c|cccccccc|cccccccc}
\toprule
Dataset$\rightarrow$     & \multicolumn{8}{c|}{CIFAR-10}                                                                                                                                                                                    & \multicolumn{8}{c}{GTSRB}                                                                                                                                                                                       \\ \hline
Attack$\rightarrow$      & \multicolumn{2}{c|}{BadNets}                            & \multicolumn{2}{c|}{Blended}                            & \multicolumn{2}{c|}{Label-consistent}                   & \multicolumn{2}{c|}{WaNet}         & \multicolumn{2}{c|}{BadNets}                            & \multicolumn{2}{c|}{Blended}                            & \multicolumn{2}{c|}{Label-consistent}                   & \multicolumn{2}{c}{WaNet}         \\ \hline
\tabincell{c}{Metric$\rightarrow$\\Defense$\downarrow$}      & \multicolumn{1}{c|}{BA}    & \multicolumn{1}{c|}{ASR}   & \multicolumn{1}{c|}{BA}    & \multicolumn{1}{c|}{ASR}   & \multicolumn{1}{c|}{BA}    & \multicolumn{1}{c|}{ASR}   & \multicolumn{1}{c|}{BA}    & ASR   & \multicolumn{1}{c|}{BA}    & \multicolumn{1}{c|}{ASR}   & \multicolumn{1}{c|}{BA}    & \multicolumn{1}{c|}{ASR}   & \multicolumn{1}{c|}{BA}    & \multicolumn{1}{c|}{ASR}   & \multicolumn{1}{c|}{BA}    & ASR   \\ \hline
No Defense  & \multicolumn{1}{c|}{92.02} & \multicolumn{1}{c|}{\red{96.96}} & \multicolumn{1}{c|}{90.89} & \multicolumn{1}{c|}{\red{83.11}} & \multicolumn{1}{c|}{92.17} & \multicolumn{1}{c|}{\red{100}}   & \multicolumn{1}{c|}{91.48} & \red{96.82} & \multicolumn{1}{c|}{98.25} & \multicolumn{1}{c|}{\red{95.38}} & \multicolumn{1}{c|}{98.65} & \multicolumn{1}{c|}{\red{87.48}} & \multicolumn{1}{c|}{96.83} & \multicolumn{1}{c|}{66.40} & \multicolumn{1}{c|}{96.37} & 77.53 \\ \hline
FT-FC       & \multicolumn{1}{c|}{\textbf{92.08}} & \multicolumn{1}{c|}{\red{97.41}} & \multicolumn{1}{c|}{\underline{91.34}} & \multicolumn{1}{c|}{\red{83.17}} & \multicolumn{1}{c|}{\textbf{92.42}} & \multicolumn{1}{c|}{\red{100}}   & \multicolumn{1}{c|}{\textbf{91.26}} & \red{98.26} & \multicolumn{1}{c|}{98.62} & \multicolumn{1}{c|}{\red{95.63}} & \multicolumn{1}{c|}{91.34} & \multicolumn{1}{c|}{\red{83.17}} & \multicolumn{1}{c|}{97.62} & \multicolumn{1}{c|}{77.19} & \multicolumn{1}{c|}{97.35} & 67.93 \\
FT-All      & \multicolumn{1}{c|}{\underline{91.66}} & \multicolumn{1}{c|}{\red{96.31}} & \multicolumn{1}{c|}{\textbf{91.76}} & \multicolumn{1}{c|}{0.81}  & \multicolumn{1}{c|}{\underline{91.80}} & \multicolumn{1}{c|}{\red{86.79}} & \multicolumn{1}{c|}{\underline{90.98}} & 62.18 & \multicolumn{1}{c|}{\textbf{99.78}} & \multicolumn{1}{c|}{22.11} & \multicolumn{1}{c|}{\textbf{99.62}} & \multicolumn{1}{c|}{2.74}  & \multicolumn{1}{c|}{\textbf{99.68}} & \multicolumn{1}{c|}{57.22} & \multicolumn{1}{c|}{\textbf{99.60}} & 11.32 \\ \hline
MCR         & \multicolumn{1}{c|}{84.00} & \multicolumn{1}{c|}{0.80}  & \multicolumn{1}{c|}{82.51} & \multicolumn{1}{c|}{0.71}  & \multicolumn{1}{c|}{84.13} & \multicolumn{1}{c|}{\underline{6.15}}  & \multicolumn{1}{c|}{87.77} & 14.28 & \multicolumn{1}{c|}{98.50} & \multicolumn{1}{c|}{\textbf{0.16}}  & \multicolumn{1}{c|}{98.38} & \multicolumn{1}{c|}{\textbf{0.11}}  & \multicolumn{1}{c|}{95.85} & \multicolumn{1}{c|}{\underline{0.03}}  & \multicolumn{1}{c|}{96.78} & \textbf{0.00}  \\
NAD         & \multicolumn{1}{c|}{90.66} & \multicolumn{1}{c|}{0.94}  & \multicolumn{1}{c|}{90.00} & \multicolumn{1}{c|}{0.86}  & \multicolumn{1}{c|}{90.20} & \multicolumn{1}{c|}{17.72} & \multicolumn{1}{c|}{90.09} & 11.34 & \multicolumn{1}{c|}{96.91} & \multicolumn{1}{c|}{0.46}  & \multicolumn{1}{c|}{97.09} & \multicolumn{1}{c|}{\underline{0.21}}  & \multicolumn{1}{c|}{96.58} & \multicolumn{1}{c|}{\textbf{0.00}}  & \multicolumn{1}{c|}{97.12} & 37.88 \\ \hline
SSFT        & \multicolumn{1}{c|}{89.08} & \multicolumn{1}{c|}{\textbf{0.31}}  & \multicolumn{1}{c|}{88.64} & \multicolumn{1}{c|}{\textbf{0.31}}  & \multicolumn{1}{c|}{89.86} & \multicolumn{1}{c|}{\textbf{4.63}}  & \multicolumn{1}{c|}{88.76} & \textbf{0.40}  & \multicolumn{1}{c|}{99.62} & \multicolumn{1}{c|}{\underline{0.33}}  & \multicolumn{1}{c|}{\underline{99.46}} & \multicolumn{1}{c|}{0.24}  & \multicolumn{1}{c|}{\underline{99.60}} & \multicolumn{1}{c|}{6.00}  & \multicolumn{1}{c|}{99.43} & \underline{9.69}  \\ 
SSFT*       & \multicolumn{1}{c|}{90.04} & \multicolumn{1}{c|}{\underline{0.37}}  & \multicolumn{1}{c|}{89.22} & \multicolumn{1}{c|}{\underline{0.32}}  & \multicolumn{1}{c|}{90.26} & \multicolumn{1}{c|}{6.44}  & \multicolumn{1}{c|}{88.80} & \underline{1.38}  & \multicolumn{1}{c|}{\underline{99.68}} & \multicolumn{1}{c|}{0.82}  & \multicolumn{1}{c|}{99.45} & \multicolumn{1}{c|}{0.41}  & \multicolumn{1}{c|}{\underline{99.60}} & \multicolumn{1}{c|}{6.06}  & \multicolumn{1}{c|}{\underline{99.44}} & 12.83 \\ \bottomrule
\end{tabular}
}
\begin{tablenotes}
\footnotesize
\item[1]SSFT: shortcut suppression with fine-tuning under the standard mode ($i.e.$, the decay factor $\gamma$ is set to 0).
\item[2]SSFT*: shortcut suppression with fine-tuning under the optimized mode ($i.e.$, with optimal decay factor).
\end{tablenotes}
\end{threeparttable}
\end{table*}

\begin{figure}[!t]
    \centering
    \includegraphics[width=0.47\textwidth]{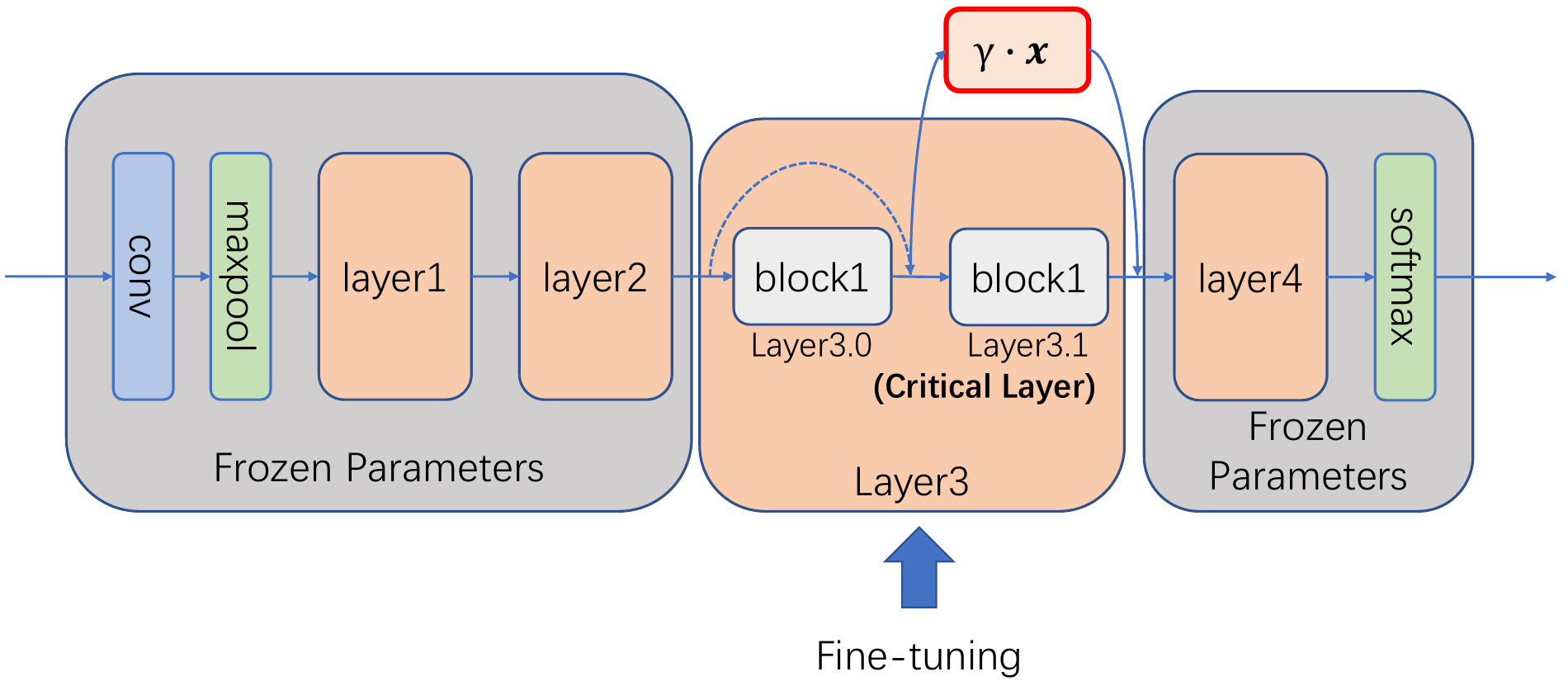}
    \vspace{-1em}
    \caption{The main pipeline of our method. In the first stage, we select the critical layer based on surrogate poisoned dataset. After that, we suppress its skip connection by multiplying a small decay factor. We fine-tune the critical layer at the end. } 
    \label{fig:the whole pipeline}
    \vspace{-1em}
\end{figure}

\subsection{Shortcut Suppression with Fine-Tuning (SSFT)}

\noindent\textbf{Overview.} In general, our SSFT consists of three main stages (as shown in Figure \ref{fig:the whole pipeline}), including \textbf{1)} selecting critical layer(s) \textbf{2)} suppressing shortcut(s), and \textbf{3)} fine-tuning critical layer(s).  



\vspace{0.3em}
\noindent\textbf{Selecting Critical Layer.} As demonstrated in the previous section, different attacks tend to adopt similar critical layers for backdoor injection. Motivated by this observation, we propose to adopt a surrogate poisoned dataset to select the critical layer. Specifically, we retain the obtained third-party model with the surrogate dataset, based on which to select the layer by gradually decreasing the introduced decay factor of the skip connection in each layer. Layers having the sharpest ASR decreases are our critical layers.

\vspace{0.3em}
\noindent\textbf{Shortcut Suppression.} 
Once critical layers are selected, we suppress its skip connection of the given third-party suspicious model by using a small decay factor $\gamma$ during the inference process. Specifically, we introduce two suppression modes, including \emph{standard mode} and \emph{optimized mode}. In the standard mode, we simply assign $\gamma$ as 0, $i.e.$, remove the skip connection. We adopt the optimal $\gamma$ selected in the process of critical layer selection in our optimal mode. We notice that our standard mode is different from using models without skip connections since we only remove the skip connection contained in the critical layer (instead of in all layers).

\begin{table*}
\centering
\vspace{-2em}
\caption{The ablation study (\%) of our method. }
\vspace{-0.8em}
\label{table:ablation}
\scalebox{0.78}{
\begin{tabular}{c|cccccccc|cccccccc}
\toprule
Dataset$\rightarrow$     & \multicolumn{8}{c|}{CIFAR-10}                                                                                                                                                                                    & \multicolumn{8}{c}{GTSRB}                                                                                                                                                                                       \\ \hline
Attack$\rightarrow$      & \multicolumn{2}{c|}{BadNets}                            & \multicolumn{2}{c|}{Blended}                            & \multicolumn{2}{c|}{Label-consistent}                   & \multicolumn{2}{c|}{WaNet}         & \multicolumn{2}{c|}{BadNets}                            & \multicolumn{2}{c|}{Blended}                            & \multicolumn{2}{c|}{Label-consistent}                   & \multicolumn{2}{c}{WaNet}         \\ \hline
\tabincell{c}{Metric$\rightarrow$\\Defense$\downarrow$}      & \multicolumn{1}{c|}{BA}    & \multicolumn{1}{c|}{ASR}   & \multicolumn{1}{c|}{BA}    & \multicolumn{1}{c|}{ASR}   & \multicolumn{1}{c|}{BA}    & \multicolumn{1}{c|}{ASR}   & \multicolumn{1}{c|}{BA}    & ASR   & \multicolumn{1}{c|}{BA}    & \multicolumn{1}{c|}{ASR}   & \multicolumn{1}{c|}{BA}    & \multicolumn{1}{c|}{ASR}   & \multicolumn{1}{c|}{BA}    & \multicolumn{1}{c|}{ASR}   & \multicolumn{1}{c|}{BA}    & ASR   \\ \hline
No Defense  & \multicolumn{1}{c|}{92.02} & \multicolumn{1}{c|}{96.96} & \multicolumn{1}{c|}{90.89} & \multicolumn{1}{c|}{83.11} & \multicolumn{1}{c|}{92.17} & \multicolumn{1}{c|}{100}   & \multicolumn{1}{c|}{91.48} & 96.82 & \multicolumn{1}{c|}{98.25} & \multicolumn{1}{c|}{95.38} & \multicolumn{1}{c|}{98.65} & \multicolumn{1}{c|}{87.48} & \multicolumn{1}{c|}{96.83} & \multicolumn{1}{c|}{66.40} & \multicolumn{1}{c|}{96.37} & 77.53 \\ \hline
SS          & \multicolumn{1}{c|}{14.30} & \multicolumn{1}{c|}{0.00}  & \multicolumn{1}{c|}{21.57} & \multicolumn{1}{c|}{0.00}  & \multicolumn{1}{c|}{18.65} & \multicolumn{1}{c|}{0.00}  & \multicolumn{1}{c|}{12.78} & 0.00  & \multicolumn{1}{c|}{97.61} & \multicolumn{1}{c|}{0.65}  & \multicolumn{1}{c|}{87.99} & \multicolumn{1}{c|}{25.13} & \multicolumn{1}{c|}{88.44} & \multicolumn{1}{c|}{6.39}  & \multicolumn{1}{c|}{22.41} & 0.00  \\ 
FT-Critical & \multicolumn{1}{c|}{91.82} & \multicolumn{1}{c|}{96.73} & \multicolumn{1}{c|}{91.74} & \multicolumn{1}{c|}{55.86} & \multicolumn{1}{c|}{92.12} & \multicolumn{1}{c|}{99.30} & \multicolumn{1}{c|}{91.06} & 93.46 & \multicolumn{1}{c|}{98.32} & \multicolumn{1}{c|}{95.10} & \multicolumn{1}{c|}{99.47} & \multicolumn{1}{c|}{59.68} & \multicolumn{1}{c|}{99.53} & \multicolumn{1}{c|}{88.42} & \multicolumn{1}{c|}{99.23} & 45.09 \\ 
SSFT        & \multicolumn{1}{c|}{89.08} & \multicolumn{1}{c|}{0.31}  & \multicolumn{1}{c|}{88.64} & \multicolumn{1}{c|}{0.31}  & \multicolumn{1}{c|}{89.86} & \multicolumn{1}{c|}{4.63}  & \multicolumn{1}{c|}{88.76} & 0.40  & \multicolumn{1}{c|}{99.62} & \multicolumn{1}{c|}{0.33}  & \multicolumn{1}{c|}{99.46} & \multicolumn{1}{c|}{0.24}  & \multicolumn{1}{c|}{99.60} & \multicolumn{1}{c|}{6.00}  & \multicolumn{1}{c|}{99.43} & 9.69 \\ \bottomrule
\end{tabular}
}
\vspace{0.5em}
\end{table*}

\vspace{0.3em}
\noindent\textbf{Fine-tuning Critical Layer.} 
Shortcut suppression can significantly reduce attack success rate. However, as shown in Figure \ref{fig:experiments in inference stage}, this process also has negative effects on benign accuracy (BA). Especially under the standard mode, the BA may even drop to 0. This phenomenon is expected since the prediction of benign samples also relies heavily on the skip connection. Accordingly, we propose to fine-tune selected critical layer(s) based on local benign samples. This stage can significantly increase BA and further reduce ASR.

\section{Experiments}

\subsection{Settings}
\textbf{Dataset and Model.} 
We conduct experiments on two classical benchmark datasets, including CIFAR-10 and GTSRB datasets, with ResNet-18 \cite{he2016deep}. 


\vspace{0.3em}
\noindent\textbf{Attack Baselines.}
We choose BadNets \cite{gu2019badnets}, backdoor attack with blended strategy (dubbed `Blended') \cite{chen2017targeted}, label-consistent attack with adversarial perturbations (dubbed `Label-Consistent') \cite{turner2019label}, and WaNet \cite{nguyen2021wanet} as our attack baselines. They are the representatives of patch-based visible attacks and patch-based invisible attacks, clean-label attacks, and non-patch-based invisible attacks, respectively.


\vspace{0.3em}
\noindent\textbf{Defense Baselines.}
We compare our methods with two state-of-the-art model-repairing-based backdoor defenses, including neural attention distillation (NAD) \cite{li2020neural} and mode connectivity repair (MCR) \cite{zhao2020bridging}. We also adopt fine-tuning on fully-connected layers (dubbed `FT-FC'), all layers (dubbed `FT-All'), and standard model training (dubbed `No Defense') as other important baselines for our discussions. 


\begin{figure*}[ht]
\centering
\begin{minipage}[t]{0.32\linewidth}
\centering
\includegraphics[width=\textwidth]{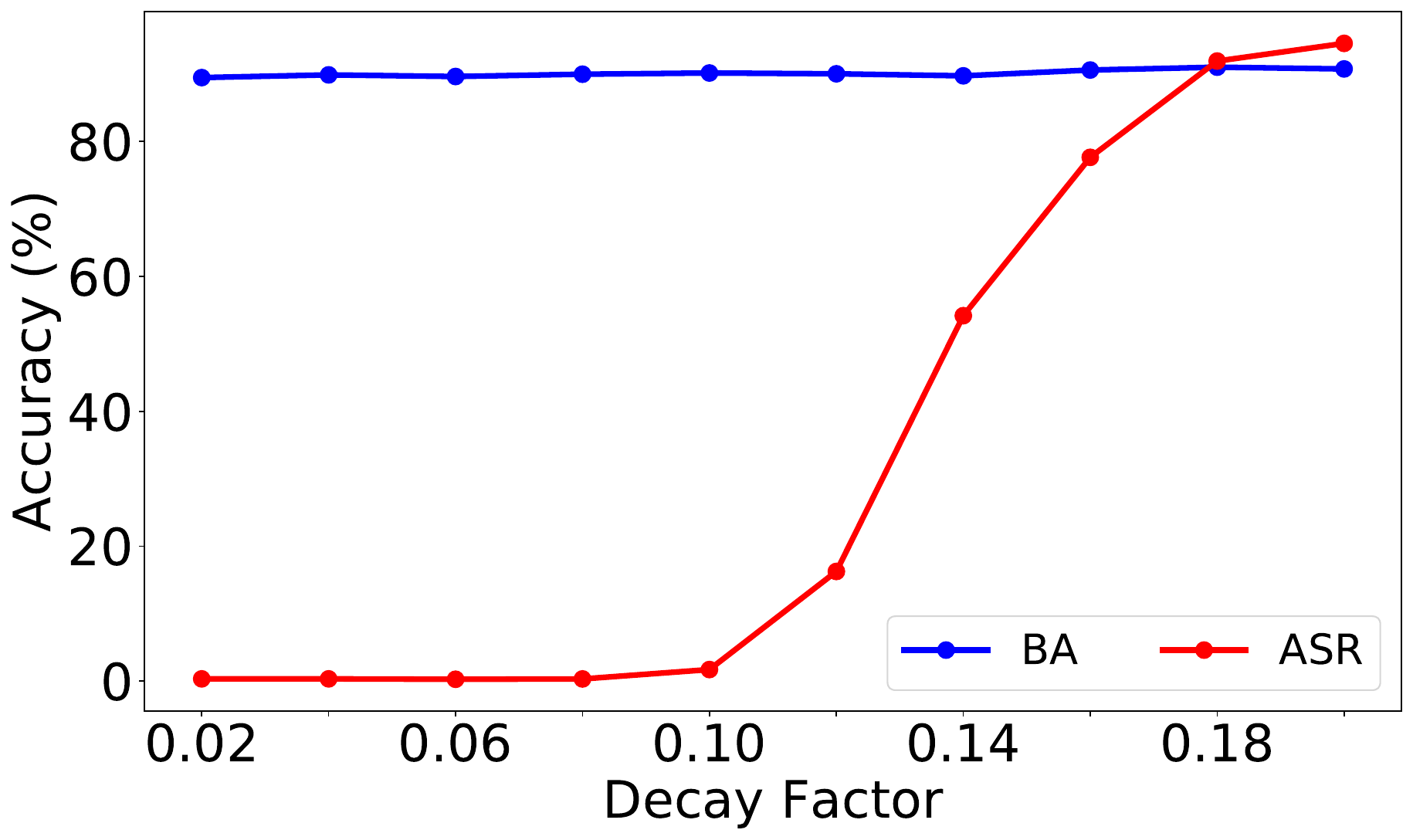}
\vspace{-1.3em}
\caption{Effects of the decay factors.}
\label{fig:weight}
\end{minipage}\hspace{0.3em}
\begin{minipage}[t]{0.32\linewidth}
\centering
\includegraphics[width=\textwidth]{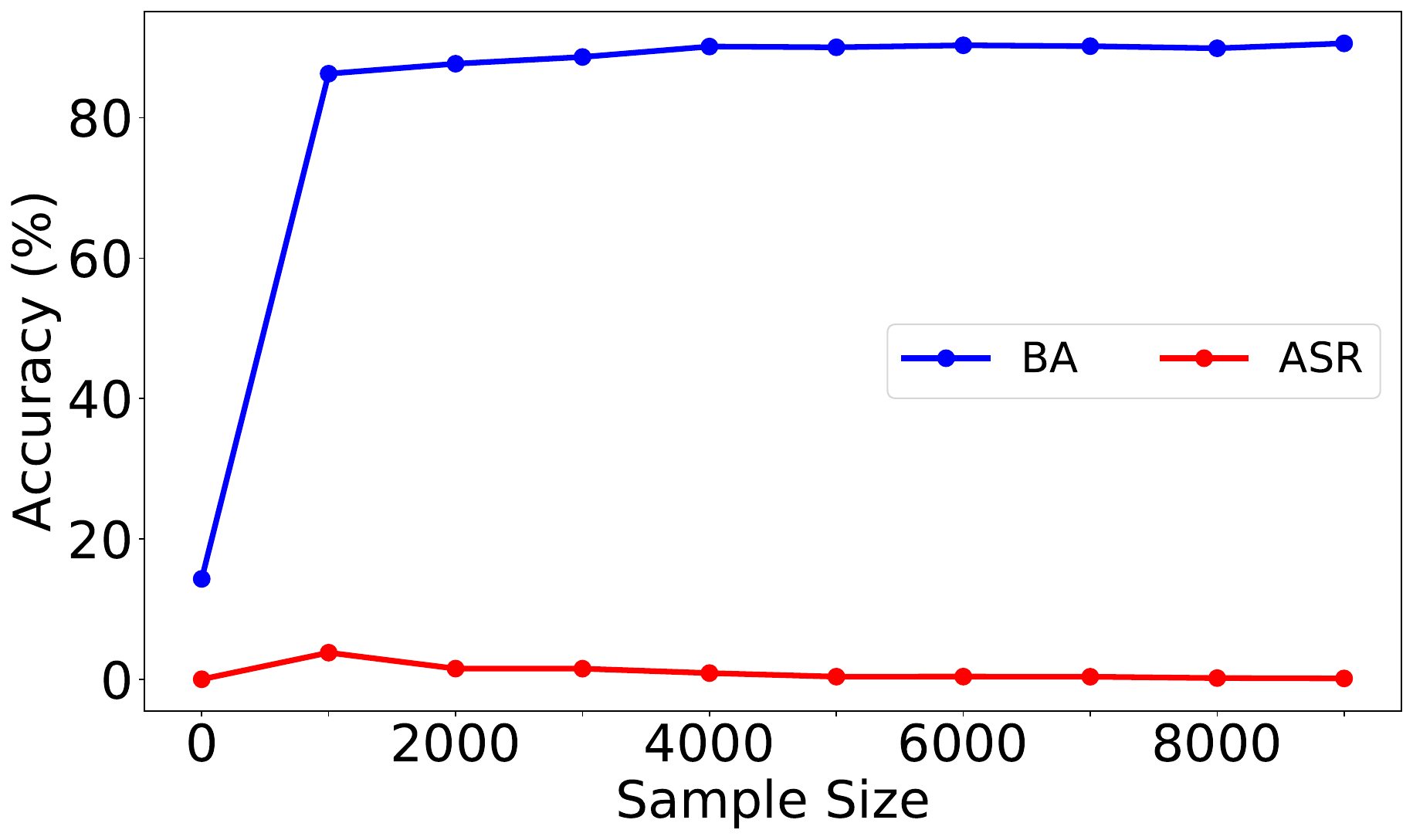}
\vspace{-1.3em}
\caption{Effects of the sample size.}
\label{fig:sample size}
\end{minipage}\hspace{0.3em}
\begin{minipage}[t]{0.32\linewidth}
\centering
\includegraphics[width=\textwidth]{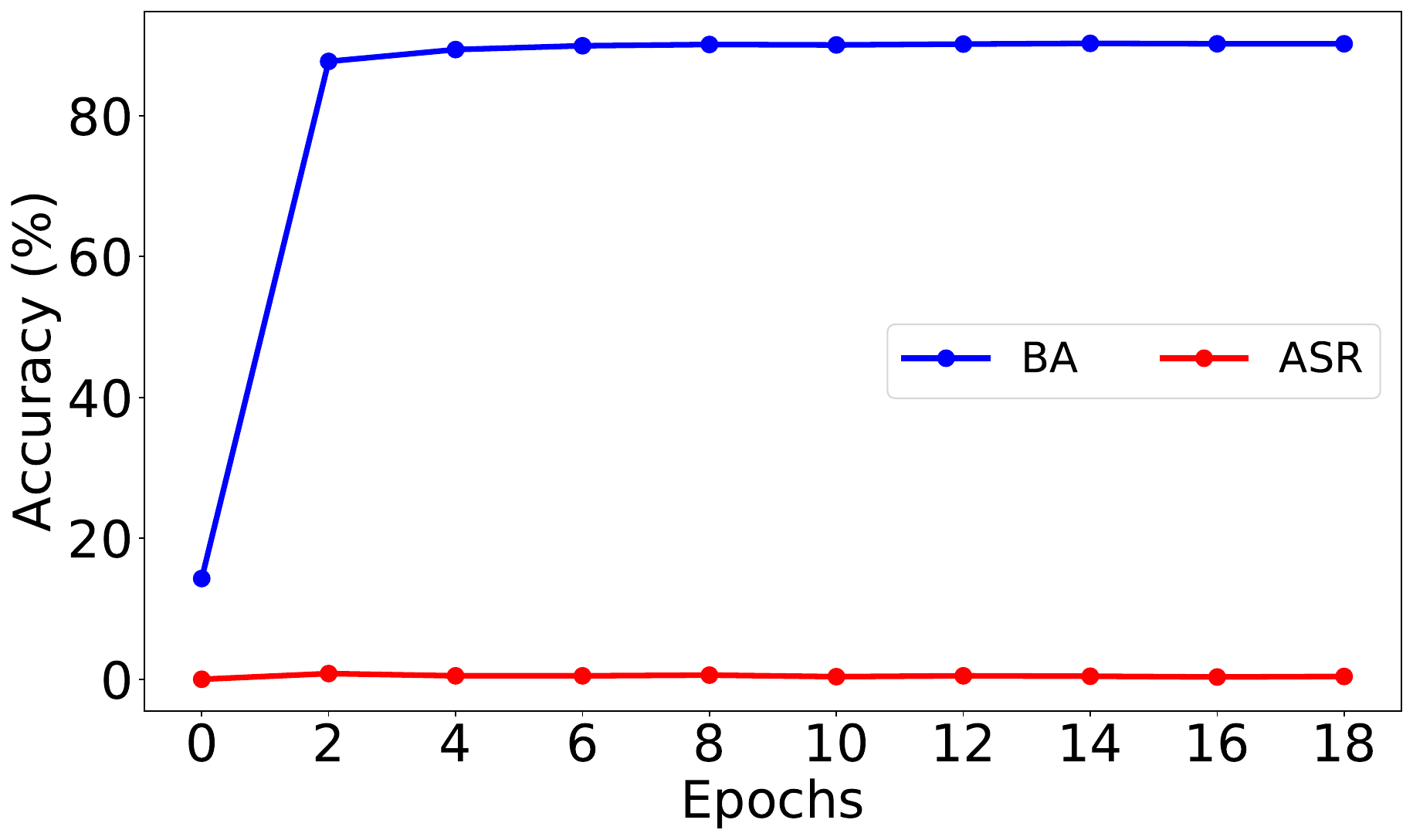}
\vspace{-1.3em}
\caption{Effects of the tuning epoch.}
\label{fig:epochs}
\end{minipage}\hspace{0.3em}
\end{figure*}


\vspace{0.3em}
\noindent\textbf{Attack Setup.}
For BadNets and Blended, we set a $3\times 3$ black-white patch as the trigger pattern on both CIFAR-10 and GTSRB. We implement label-consistent attack and WaNet based on the \texttt{BackdoorBox} \cite{li2023backdoorbox} with their default settings. We set the poisoning rate as 5\% for all attacks to generate the poisoned datasets. The target label is set to 1 for all attacks.

\vspace{0.3em}
\noindent\textbf{Defense Setup.}
In all fine-tuning-based methods, we set the learning rate as 0.01 and use 10\% benign samples. We fine-tune the model 10 epochs in total. For NAD and MCR, we also use 10\% benign training samples which are also exploited in our defense. All baseline defenses are also implemented based on the \texttt{BackdoorBox}.

\vspace{0.3em}
\noindent\textbf{Evaluation Metric.}
Following the classical settings used in existing backdoor defenses, we adopt the attack success rate (ASR) and benign accuracy (BA) to measure the effectiveness of all methods. In general, the lower the ASR and the higher the BA, the better the defense.

\subsection{Main Results}
As shown in Table \ref{table:experiments results}, both FT-FC and FT-All fail in many cases with a high attack success rate (ASR), although they usually preserve the highest benign accuracy (BA). In contrast, both MCR and NAD reach relatively low ASR values in general, whereas having low BA values. Different from previous baseline methods, our SSFT yields a more balanced performance. In other words, our defense significantly reduces ASR while preserving high BA. We notice that SSFT may also have better performance compared to SSFT*. It is mostly because the optimal decay factor is selected based on surrogate poisoned samples which may not be consistent with those used by the adversaries. We will further explore how to better select the decay factor in our future work.




\vspace{-1em}
\subsection{Discussion}

\noindent \textbf{Ablation Study.}
There are two key strategies in our SSFT defense, including shortcut suppression and model fine-tuning. In this part, we discuss their effects. As shown in Table \ref{table:ablation}, shortcut suppression (SS) significantly reduces the attack success rate (ASR). However, it also results in low benign accuracy (BA). Fine-tuning the critical layer preserves high BA while having minor effects in decreasing ASR. In contrast, our SSFT is effective in maintaining BA and removing hidden backdoors. These results verify that both of these strategies are indispensable parts of our method.



\vspace{0.3em}
\noindent \textbf{Effects of the Decay Factor.}
In this part, we change the decay factor $\gamma$ used in our SSFT while preserving other settings. As is shown in Figure \ref{fig:weight}, both benign accuracy and attack success rate are stable when using relatively small $\gamma$ ($i.e.$, $<0.1$). In other words, our method is not very sensitive to the selection of $\gamma$ to some extent. However, we need to notice that using a large decay factor will reduce defense effectiveness as it will degenerate our SSFT back to the model fine-tuning.


\vspace{0.3em}
\noindent \textbf{Effects of the Sample Size.}
Recall that our SSFT method needs some benign local samples for fine-tuning. Here we discuss how the sample size influences our defense. As shown in Figure \ref{fig:sample size}, the BA increases while the ASR decreases with the increase in sample size. In particular, the SSFT could achieve promising performance even with a small sample size. These results reflect the effectiveness of our defense.


\vspace{0.3em}
\noindent \textbf{Effects of the Tuning Epoch.}
Here we explore the influence of tuning epoch on our SSFT defense. As shown in Figure \ref{fig:epochs}, similar to the effects of the sample size, the BA increases while the ASR decreases with the increase in the epoch. Our method can achieve good performance with a few epochs ($e.g.$, 4), showing high efficiency.

\section{Conclusions}
In this paper, we explored the backdoor mechanism from the aspect of model structures. We revealed that backdoor threats have an underlying correlation to the skip connection. Specifically, we demonstrated that suppressing the skip connection in critical layers can significantly reduce the attack success rate. Based on these findings, we propose a simple yet effective backdoor defense, dubbed shortcut suppression with fine-tuning (SSFT), to remove hidden backdoors contained in the given suspicious model. 
Extensive experiments on benchmark datasets verified the effectiveness of our defense. 

\section*{Acknowledgement}
\vspace{-0.4em}
This work is supported in part by the National Natural Science Foundation of China under Grant 62771248, Shenzhen Science and Technology Program (JCYJ20220818101012025), the PCNL KEY project (PCL2021A07), and Research Center for Computer Network (Shenzhen) Ministry of Education.
\vfill\pagebreak

\label{sec:ref}

\bibliographystyle{IEEEbib}
\bibliography{ms}

\end{document}